\documentclass[10pt,twocolumn,letterpaper]{article}

\usepackage{ICCV2025-Author-Kit-Feb/iccv}              

\usepackage[accsupp]{axessibility}  

\usepackage[acronym,nomain,nonumberlist]{glossaries}
\makeglossaries

\newacronym{dl}{DL}{Deep Learning}
\newacronym{ai}{AI}{Artificial Intelligence}
\newacronym{cw}{CW}{Concept Whitening}
\newacronym{clarc}{ClArC}{Class Artifact Compensation}
\newacronym{pcbm}{{PCBM}}{Post-hoc Concept Bottleneck Model}
\newacronym{lcr}{LCR}{Latent Concept Representation}
\newacronym{cav}{CAV}{Concept Activation Vector}
\newacronym{rcv}{RCV}{Regression Concept Vector}
\newacronym{car}{CAR}{Concept Activation Region}
\newacronym{lcrreg}{LCRReg}{LCR Regularization}
\newacronym{svm}{SVM}{Support Vector Machine}
\newacronym{db}{DB}{Decision Boundary}
\newacronym{nn}{NN}{Neural Network}
\newacronym{ba}{BA}{Balanced Accuracy}
\newacronym{dr}{DR}{Diabetic Retinopathy}
\newacronym{ood}{OOD}{out-of-distribution}
\newacronym{pcbmh}{PCBM-h}{Hybrid Post-hoc Concept Bottleneck Model}
\newacronym{mtl}{MTL}{Multi-Task Learning}
\newacronym{ce}{CE}{Cross-Entropy}

\definecolor{iccvblue}{rgb}{0.21,0.49,0.74}
\usepackage[pagebackref,breaklinks,colorlinks,allcolors=iccvblue]{hyperref}

\def\paperID{8} 
\def\confName{ICCV}
\def\confYear{2025}

\title{In-hoc Concept Representations to Regularise Deep Learning in Medical Imaging}

\author{
  Valentina Corbetta\thanks{Equal contribution} \textsuperscript{1,2,3} \quad
  Floris Six Dijkstra\footnotemark[1] \textsuperscript{1,4} \quad
  Regina Beets-Tan\textsuperscript{1,3} \quad
  Hoel Kervadec\textsuperscript{4,5} \\[0.5ex]
  Kristoffer Wickstrøm\textsuperscript{6} \quad
  Wilson Silva\textsuperscript{1,2} \\[1ex]
  \small
  \textsuperscript{1}The Netherlands Cancer Institute, The Netherlands \quad
  \textsuperscript{2}Utrecht University, The Netherlands \quad
  \textsuperscript{3}Maastricht University, The Netherlands \\[0.5ex]
  \small
  \textsuperscript{4}University of Amsterdam, The Netherlands \quad
  \textsuperscript{5}Amsterdam UMC, The Netherlands \quad
  \textsuperscript{6}UiT The Arctic University of Norway, Norway \\
  {\tt\small \{v.corbetta, r.beetstan\}@nki.nl,
            floris.six.dijkstra@student.uva.nl,
            h.t.g.kervadec@uva.nl,}\\[-0.3ex]
  {\tt\small kristoffer.k.wickstrom@uit.no,
            w.j.dossantossilva@uu.nl}
}

\begin{document}
\maketitle
\begin{abstract}
Deep learning models in medical imaging often achieve strong in-distribution performance but struggle to generalise under distribution shifts, frequently relying on spurious correlations instead of clinically meaningful features. We introduce LCRReg, a novel regularisation approach that leverages Latent Concept Representations (LCRs) (e.g., Concept Activation Vectors (CAVs)) to guide models toward semantically grounded representations. LCRReg requires no concept labels in the main training set and instead uses a small auxiliary dataset to synthesise high-quality, disentangled concept examples. We extract LCRs for predefined relevant features, and incorporate a regularisation term that guides a Convolutional Neural Network (CNN) to activate within latent subspaces associated with those concepts. We evaluate LCRReg across synthetic and real-world medical tasks. On a controlled toy dataset, it significantly improves robustness to injected spurious correlations and remains effective even in multi-concept and multiclass settings. On the diabetic retinopathy binary classification task, LCRReg enhances performance under both synthetic spurious perturbations and out-of-distribution (OOD) generalisation. Compared to baselines, including multitask learning, linear probing, and post-hoc concept-based models, LCRReg offers a lightweight, architecture-agnostic strategy for improving model robustness without requiring dense concept supervision. Code is available at this \href{https://github.com/Trustworthy-AI-UU-NKI/lcr\_regularization}{link}. 
\end{abstract}    
\section{Introduction}
\label{sec:intro}

\gls{dl} has driven impressive advance in computer vision and medical image analysis, achieving expert-level performance on tasks such as object classification, lesion detection, and disease diagnosis~\cite{choy2023systematic,huang2023self}. However, real-world deployment of these models remains limited due to their lack of robustness, particularly in the presence of domain shifts and spurious correlations~\cite{juodelyte2024source,liu2022learning}. In clinical practice, data often comes from varied acquisition settings, institutions, and populations, which can cause critical shifts in distribution. At the same time, training datasets may contain spurious correlations, i.e., features that are statistically predictive but clinically irrelevant, that models can rely on, leading to degraded performance when those correlations break~\cite{sun2023right}. 



Improving robustness to such phenomena is essential for safe and reliable medical \gls{ai}~\cite{litjens2017survey,shorten2019survey}. Traditional techniques such as data augmentation, domain adaptation, and adversarial training provide partial solutions, but have critical limitations. Data augmentation may not sufficiently simulate the full range of distributional variability encountered at test time~\cite{taori2020measuring}. Domain adaptation approaches often rely on access to unlabelled or partially labelled target-domain data, which is not always available in real-world clinical deployment settings~\cite{li2025semi}. Furthermore, adversarial training, while helpful for local robustness, does not distinguish between spurious and meaningful features, and may over-regularise and harm performance on unperturbed data, failing to suppress spurious correlations that are invariant to small perturbations~\cite{anders2020findingremovingcleverhans}. As noted in recent surveys~\cite{liang2025comprehensive,yoon2023domain}, these assumptions severely constrain the generalisability and practical utility of these methods in healthcare applications.

\begin{figure*}[ht]
    \centering
    \includegraphics[width=\textwidth]{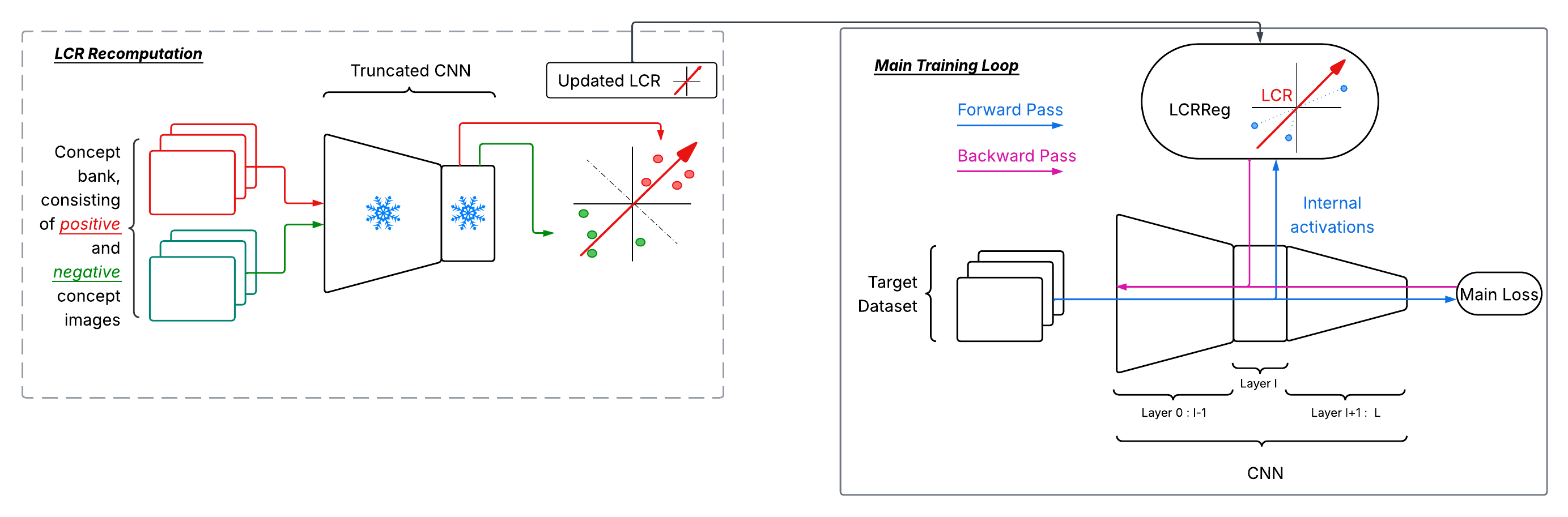}
    \caption{Illustration of the proposed LCRReg. The main training loop optimises a CNN using the $\mathcal{L}_{Main Loss}$ and a weighted Latent Concept Representation (LCR) Loss ($\mathcal{L}_{LCRReg}$). At the start of each recomputation interval $I_{rec}$, LCR recomputation is executed: concept images from the concept bank are passed through the truncated CNN to compute updated LCRs based on intermediate activations. These LCRs are then used for regularisation in subsequent training steps.}
    \label{fig:lcrreg_overview}
\end{figure*}

An emerging alternative strategy is to constrain models to rely on semantically meaningful features, such as expert-defined medical concepts. The motivation is that models grounded in medically relevant features are less likely to rely on spurious correlations and more likely to generalise across domains. Several recent works support this direction. For example, \gls{cw} aligns latent features with axes corresponding to predefined concepts~\cite{chen2020concept}, and MICA integrates multi-level concept supervision to enhance robustness and interpretability~\cite{bie2024mica}. \gls{clarc} suppresses spurious dependencies by projecting representations away from known artifact directions~\cite{anders2020findingremovingcleverhans}. By contrast, \gls{pcbm} uses \glspl{cav} for test-time interpretability, and its variant \gls{pcbmh} then applies a separate residual-fitting step to recover any predictive signal the \glspl{cav} themselves miss~\cite{yuksekgonul2022post}. Notably, these methods usually require dense concept supervision throughout the training dataset or rely on modifying the model architecture, limiting their scalability.


In this work, we pursue a general and scalable version of this concept-regularisation strategy. We introduce \gls{lcrreg}, which leverages \glspl{lcr}, i.e., post-hoc vectors that describe directions or regions in the model's activation space corresponding to clinically meaningful concepts. Examples include \glspl{cav}~\cite{kim2018interpretability}, pattern-\glspl{cav}~\cite{pahde2024navigatingneuralspacerevisiting}, \glspl{rcv}~\cite{graziani2018regression}, and \glspl{car}~\cite{crabbe2022concept}. Unlike prior works that use \glspl{lcr} purely for explanation, we incorporate them directly into training as a form of semantic regularisation. Through a regularisation loss that encourages alignment with concept subspaces or separation from spurious ones, we guide the model to encode more robust, clinically grounded representations.



Our contributions are the following:

\begin{itemize}
    \item Concept-based regularisation for robustness: We propose \gls{lcrreg}, a method that improves robustness to spurious correlations and \gls{ood} shifts by aligning intermediate representations with clinically meaningful concept directions. 
    \item Training without paired concept-image labels: Our method only requires concept annotations in a small auxiliary dataset and does not rely on concept supervision in the main training set, making it scalable and practical.
    \item Systematic comparison of \gls{lcr} types and losses: We evaluate multiple concept representation methods (\glspl{cav}, pattern-\glspl{cav}, \glspl{rcv}, \glspl{car} and regularisation strategies (subspace alignment, decision-boundary distance) in terms of their robustness effects.
    \item Robustness evaluation across tasks and datasets: We demonstrate improved performance under distribution shifts, including synthetic spurious correlations and \gls{ood} benchmarks, on a toy dataset, namely Elements~\cite{nicolsonexplaining}, and the medical imaging task of \gls{dr} classification.
\end{itemize}

\section{Methodology}
\label{sec:meth}

Given a \gls{nn} 
\begin{equation}
    \mathcal{N}: \mathbb{R}^{d_0} \rightarrow \mathbb{R}^{d_L},
\end{equation}

where $d_0$ is the dimensionality of the input, and $d_L$ is the number of output units (i.e., the pre‐softmax logits for each of the $d_L$ classes),
we augment its standard training loss with a regularisation term \gls{lcrreg} that aligns internal representations with directions associated with clinically meaningful concepts:

\begin{equation}
    \mathcal{L}_{total} = \beta_t \cdot \mathcal{L}_{Main Loss} + \alpha_t \cdot \mathcal{L}_{LCRReg}
\end{equation}

Here, $\mathcal{L}_{Main Loss}$ is the classification objective function, in our case the standard \gls{ce} loss, and $\beta_t$ and $\alpha_t$ respectively weigh the relative contribution of the classification loss and of the regularisation term. The concept directions are derived from an auxiliary dataset of $K$ predefined concepts, $C = \{c_{1,l}, c_{2,l}, ..., c_{K,l}\}$ and used to compute regularisation losses over layer $l \in T$, where $T$ is a set of selected target layers. Figure~\ref{fig:lcrreg_overview} depicts an overview of the proposed approach. 

In this Section, we present the components of our method in detail, including:

\begin{itemize}
    \item The process for constructing the concept dataset $C$.
    \item The different types of \glspl{lcr} we evaluate.
    \item The alternative formulations of the regularisation loss.
    \item The training procedures used to integrate regularisation into learning.
\end{itemize}

\begin{figure}[t]
    \centering
    \includegraphics[width=\columnwidth]
    {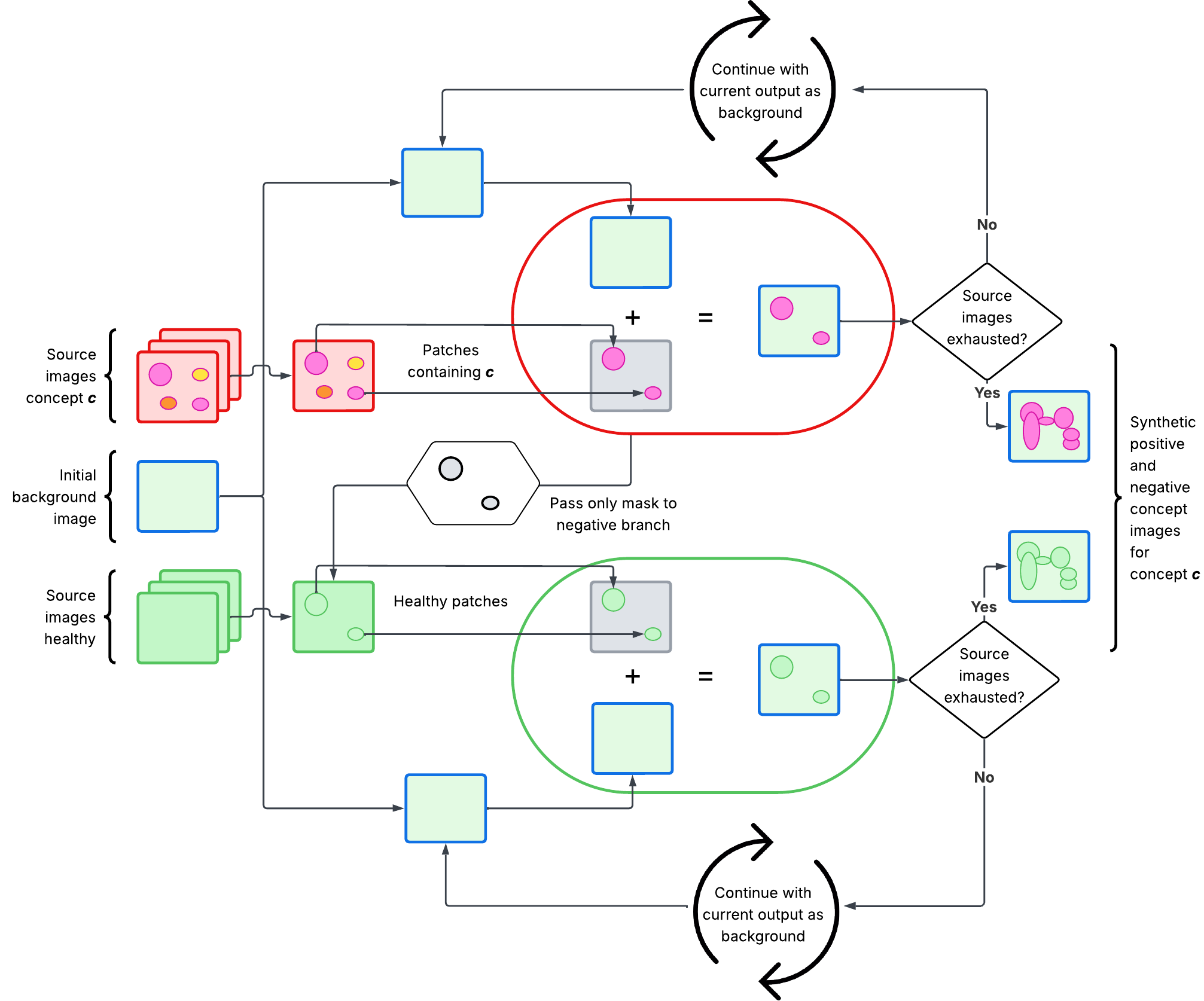}
    \caption{Overview of the concept creation pipeline. Positive samples are generated by pasting patches containing concept $c$ onto healthy images. Negative samples are created similarly, using only healthy tissue patches. This ensures both image types are synthetic and differ mainly in the presence of the concept.}
    \label{fig:concept_pipeline}
\end{figure}

\subsection{Concept Creation}

Creating high-quality, disentangled concept representations is a key prerequisite of our regularisation strategy. In natural image settings, such concepts are often visually distinct and easily annotated (e.g., ``striped'', ``red''). However, in medical imaging, clinically relevant features are typically subtle, domain-specific and difficult to label, requiring expert-level pixel-wise annotations. These annotations are expensive and time-consuming to obtain, making scalable concept-based training particularly challenging. Additionally, while automated concept extraction methods exist~\cite{oikarinen2023label}, their applicability in medical imaging remains limited due to the subtlety and domain specificity of relevant features~\cite{huy2025interactive}. Moreover, a key challenge lies in avoiding concept entanglement, i.e., the presence of multiple overlapping or correlated features within the same sample. 
Concept entanglement occurs when samples labelled as positive for a concept also contain features associated with other concepts (e.g., lesions often co-occur with artifacts or vessel abnormalities). If such co-occurrence is not controlled, the learnt \glspl{lcr} may reflect a mix of concepts rather than isolating the intended one, undermining both robustness and interpretability. A more detailed explanation of this problem and a visual example of concept entanglement in \gls{dr} can be found in Appendix~\ref{app:dr_concept_creation}.

To address both the scarcity of expert annotations and the risk of entangled concept signals, we design a modular concept synthesis pipeline that requires only a small number of annotated images, and systematically generates disentangled concept examples; this ensures that the only distinguishing factor between positive and negative samples is the presence or absence of the target concept.

The pipeline, illustrated in Figure~\ref{fig:concept_pipeline}, builds a synthetic concept bank through compositional editing as follows:
\begin{enumerate}
    \item We start from a set of healthy background images $\mathcal{H}$, which are free of all target concepts.
    \item For each target concept $c$, we collect a set of annotated source images $\mathcal{S}$, each paired with a binary segmentation mask $m_{c,k}$ identifying the concept's location.
    \item For each background image $h_b \in \mathcal{H}$ and for each target concept $c$, we generate synthetic concept examples through $N$ iterative modifications:
    \begin{enumerate}
        \item We initialise the image with $x_0 = h_b$.
        \item For $i = 1,..., N$:
        \begin{enumerate}
            \item We select a source image $p \in \mathcal{S}$ and we extract the concept-specific mask $m$.
            \item To generate a positive image, we paste the extracted concept patches from $p$ onto $x_{i-1}$ to create $x^+_i$.
            \item To generate a negative image, we sample a different healthy image $h_s \in \mathcal{H}: h_s \neq h_b$. We then extract a patch from a region spatially located on $h_s$ as $m$ on $p$ and we paste them onto $x_{i-1}$ to create $x^-{_i}$.  
        \end{enumerate}
    \end{enumerate}
\end{enumerate}

$N$ is a hyperparameter that controls how many different concept instances are inserted per synthetic sample. Increasing $N$ produces more densely composed examples, reinforcing the concept signal but potentially increasing visual artificiality. Conversely, a smaller $N$ yields more natural images at the cost of weaker signal. This trade-off lets us control the intensity and variability of concept presence across the dataset. 

We extend this approach to continuous-valued concepts, e.g., for regression-based \glspl{lcr} like \glspl{rcv}. In this instance, we synthesise only positive samples and assign each a scalar score based on the proportion of pixels occupied by the concept, enabling the model to learn a continuous representation of concept strength. 

Once computed, the \glspl{lcr} can be reused throughout training, making the pipeline highly annotation-efficient. While our method is designed to be general and transferable across medical tasks and modalities, its practical applicability may vary depending on concept granularity, segmentation quality, and clinical interpretability. Nonetheless, it provides a strong and scalable foundation for concept-based regularisation with minimal concept annotation. 

\subsection{Latent Concept Representations (LCRs)}

As mentioned in Section~\ref{sec:intro}, we consider four different \gls{lcr} methods for regularisation that can be plugged into our regularisation loss. 

\begin{itemize}
    \item Filter-\glspl{cav}~\cite{kim2018interpretability} leverage a binary classifier, e.g., a \gls{svm}, that learns to separate positive and negative concept examples in the activation space; the normal vector of the resulting \gls{db} defines the concept.

    \item Pattern-\glspl{cav}~\cite{pahde2024navigatingneuralspacerevisiting} define concept vectors as the direction between the class-conditional means of neural activations for concept-positive and concept-negative samples. Unlike filter-\glspl{cav}, which rely on training a classifier, pattern-\glspl{cav} compute this direction directly.

    \item \glspl{car}~\cite{crabbe2022concept} improve filter-\glspl{cav} by extending the underlying classifier to a non-linear kernel for binary classification. 

    \item \glspl{rcv}~\cite{graziani2018regression} extend filter-\glspl{cav} to use ordinal labels, by transforming the problem from binary classification to ordinal regression; an example, instead of being labelled positive or negative, is assigned a numerical value as a measure of how strongly the concept is represented in the input image. 
\end{itemize}

\subsection{Concept-based Regularisation}

We describe two variants of \gls{lcrreg}, based on the nature of the concept representation: one based on distances to concept subspaces (for vectorial \glspl{lcr}), and one based on distances from learnt \glspl{db} (for classifier-based \glspl{lcr}).

\subsubsection{Notation}
Let $\mathcal{N}_l(x) \in \mathbb{R}^{d_l}$ denote the activation of the network at layer $l \in T$. Let $\{v_{1,l}, v_{2,l}, ..., v_{K,l}\}$ be concept vectors derived from $C$ at layer $l$. These vectors define the concept subspace $\mathcal{C}_l = \textbf{span}(v_{1,l}, v_{2,l}, ..., v_{K,l}) \subset \mathbb{R}^{d_l}$. We denote the orthogonal projection of an activation $\mathcal{N}_l(x)$ onto $\mathcal{C}_l$ by $\hat{\mathcal{N}_l}(x) = \Pi_{\mathcal{C}_l}(\mathcal{N}_l(x))$. We use $D(\cdot, \cdot)$ to represent a generic distance function. 

\paragraph{Subspace-based Regularisation} 

For linear concept representations (e.g., filter-\glspl{cav}, pattern-\glspl{cav}, and \glspl{rcv}), we define the subspace-based regularisation loss as

\begin{equation}
    \mathcal{L}^{Sub}_{LCRReg} = \frac{1}{|T|} \sum_{l \in T} D(\mathcal{N}_l(x), \Pi_{\mathcal{C}_l}(\mathcal{N}_l(x))) 
\end{equation}

By adopting the cosine distance:

\begin{equation}
    D^{cos}(\mathcal{N}_l(x), \mathcal{C}_l) = 1 - \frac{\langle \mathcal{N}_l(x), \hat{\mathcal{N}_l}(x) \rangle}{\|\mathcal{N}_l(x)\|_2 \cdot \|\hat{\mathcal{N}_l}(x)\|_2 + \epsilon}
\end{equation}

with $\epsilon \approx 10^{-8}$, added for numerical stability.
The cosine distance is bounded between 0 and 1, thus encouraging stable optimisation. It promotes alignment with the concept subspace without driving activation norms toward zero, a common side effect of the alternative unbounded norm-based distance.

\subsection{Decision Boundary-based Regularisation}

When concepts are represented via learnt classifiers (e.g., filter-\glspl{cav} or \glspl{car}), we apply a different loss that penalises activations lying close to the learnt \glspl{db}.

Let $\phi_{i,l}: \mathbb{R}^{d_l} \rightarrow \mathbb{R}$ be the decision function for concept $i$ at layer $l$, which outputs the signed distance of the activation to the decision boundary. 
To model diminishing sensitivity to large distances, we apply a monotonically decreasing transformation $G: \mathbb{R} \rightarrow \mathbb{R}$. The resulting loss penalises activations closer to any \gls{db} more strongly and yields the following formulation:

\begin{equation}
    \mathcal{L}^{DB}_{LCRReg} = - \frac{1}{|T| \cdot K} \sum_{l \in T} \sum_{i=1}^K |G(\phi_{i,l}(\mathcal{N}_l(x)))|
\end{equation}

In our experiments, we use an exponential decay function $G(d) = exp(-d/c)$ with hyperparameter $c > 0$ controlling the rate of the decay, which smoothly increases the penalty for activations lying near the \gls{db}.

\subsection{Training Strategies}

Effectively combining concept-based supervision with primary task learning requires careful consideration of two key elements: first, how the relative weight of the regularisation signal evolves during training; second, how often the \glspl{lcr} are recomputed. Improper handling of either can result in underutilised supervision, unstable optimisation, or information leakage. 

\subsubsection{Regularisation Scheduling}

Rather than fixing the balance between the classification loss and regularisation throughout training, we explore different scheduling schemes for the loss weights $\alpha_t$ and $\beta_t$. These define how strongly concept signals influence learning over time. We consider three strategies:

\begin{itemize}
    \item Static regularisation: the simplest approach, where $\beta_t = 1$ and $\alpha_t$ is kept at a constant value, obtained via optimisation. To allow the network to learn useful features before regularisation begins, we optionally delay its activation by setting $\alpha_t = 0$ for $t<\tilde{t}$, where $\tilde{t}$ is a schedule hyperparameter.
    \item Dynamic regularisation: $\alpha_t$ is increased and $\beta_t$ is decreased over training. This schedule assumes that \glspl{lcr} may be more accurate at later stages, when internal features are more meaningful. When the regularisation strength $\alpha_t$ increases toward the end of training, the model relies more heavily on the concept representations. This can help override signals from spurious correlations. As a result, the model may become more robust and aligned with clinically meaningful features. However, this approach introduces additional hyperparameters, resulting in a more complex training setup compared to static regularisation. 
    \item 3-stage training: to isolate learning signals and avoid mutual interference, we adopt a staged training strategy:
    \begin{enumerate}
        \item Stage I: train the model with only the classification loss $\mathcal{L}_{Main Loss}$ ($\beta_t = 1, \alpha_t = 0$).
        \item Stage II: switch to concept regularisation only ($\beta_t=0, \alpha_t=1$).
        \item Stage III: fine-tune only layers on top of the \gls{lcr} target layers with the classification loss ($\beta_t=1, \alpha_t=0$), freezing the rest of the network.
    \end{enumerate}
\end{itemize}

The motivation behind the staged training setup is twofold:

\begin{itemize}
    \item To reduce information leakage. When concept supervision and task supervision are applied simultaneously, the model may learn to satisfy both objectives by encoding shortcuts or spurious correlations that happen to co-occur with the target concepts. \glspl{nn} are flexible enough to exploit such correlations, especially early in training when internal representations are still forming. By isolating the concept supervision phase (Stage II), we reduce gradient propagation beyond target layers, limiting the model's ability to track spurious correlations across the entire network. This encourages internal representations to align more cleanly with the intended concept subspaces, rather than entangled or confounded features. 
    \item To enable investigative comparison. The staged design also allows us to explicitly assess the value of incorporating concept supervision during feature learning: if training with \gls{lcrreg} in Stage II leads to a better final performance than applying them only post-hoc, this provides evidence that our method actively shapes internal representations in a beneficial way.  
\end{itemize}

\subsubsection{LCR Recomputation Interval} 

Unlike post-hoc interpretability settings where \glspl{lcr} are computed only once, regularisation requires integrating them into the training loop. However, since model weights evolve during training, a concept vector computed once may become outdated. We define a recomputation interval $I_{rec}$ to control how frequently \glspl{lcr} are updated. 

At one extreme, we recompute after every training epoch, which maximises alignment with the current model but introduces high computational overhead and training instability due to the moving objective. At the other extreme, recomputing only once at the beginning improves stability but may result in misaligned supervision. 

In Section~\ref{sec:exps}, we investigate the effect of the presented training strategies.

\section{Experiments} \label{sec:exps}

We evaluate our proposed method, \gls{lcrreg}, in two complementary settings: a controlled synthetic dataset, Elements, and a real-world medical imaging task, namely \gls{dr} classification. 

\subsection{Synthetic Experiments} \label{subsec:elem}

\begin{figure}[t]
    \centering
    \includegraphics[width=\columnwidth]
    {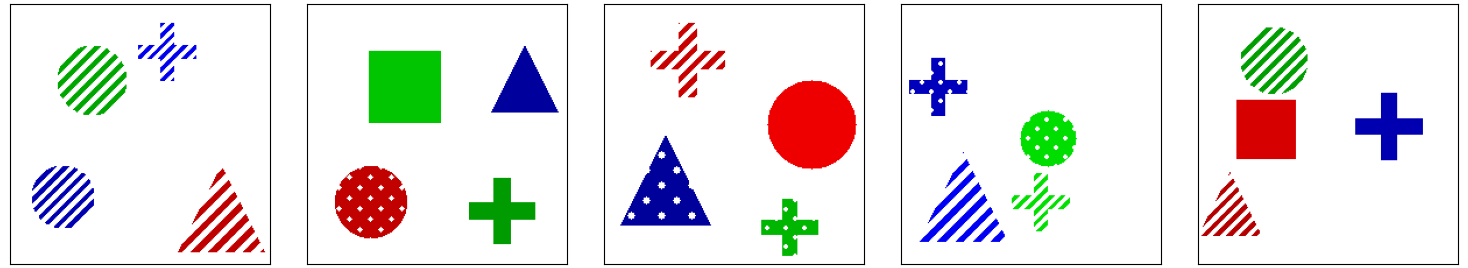}
    \caption{Sample images from the Elements dataset.}
    \label{fig:elements}
\end{figure}

We first validate \gls{lcrreg} on the synthetic Elements dataset, explicitly designed to control the presence of concepts and spurious correlations. Each image comprises distinct ``elements" described by five attributes: shape, colour, texture, location, and size (examples are reported in Figure~\ref{fig:elements}). We initially focus on a binary classification task: detecting squares. During training, squares consistently have a spurious correlation, diagonal stripes, that is absent in the test set. The dataset consists of 300 training images and 1000 images each for validation and testing. We train ResNet18~\cite{he2016deep} pretrained on ImageNet~\cite{deng2009imagenet} for 20 epochs using the Adam optimiser~\cite{kingma2014adam} with a learning rate of $10^{-5}$ and weight decay of $10^{-4}$ and report \gls{ba}. We apply \gls{lcrreg} to the $1 \times 1$ projection convolution that adjusts the residual connection in the first module of the last ResNet block.

\subsubsection{Ablation on Hyperparameter Impact}

We evaluate the impact of three key hyperparameters on the Elements dataset: \gls{lcrreg} loss weight $\alpha_t$, regularisation start epoch $\tilde{t}$, and \gls{lcr} recomputation interval $I_{rec}$. Results (\Cref{fig:hyperparameter_ablation} in Appendix~\ref{app:ablations}) show that performance improves with higher $\alpha_t$ and immediate regularisation ($\tilde{t} = 0$). Frequent \gls{lcr} recomputation introduces instability, with best results achieved by computing concepts once at the start. These findings suggest that strong, early regularisation with fixed concept vectors is effective for simple settings. 

\subsubsection{Regularisation Scheme Comparison}

We compare the three regularisation schemes introduced in Section~\ref{sec:meth}: static, dynamic, and 3-stage training. Hyperparameters are optimised via Optuna~\cite{akiba2019optuna} and evaluated statistically over 15 dataset splits; detailed results of the optimisation are reported in Appendix~\ref{app:ablations}. Figure~\ref{fig:reg_scheme_comp} in Appendix~\ref{app:ablations} shows no significant difference between static and dynamic schemes ($p = 0.76$). Static regularisation without \gls{lcr} recomputation achieves the best overall performance. 3-stage training significantly underperforms ($p < 0.0001$).

\subsubsection{Multi-concept and Multi-class settings}

\begin{figure}[t]
    \centering
    \includegraphics[width=\columnwidth]
    {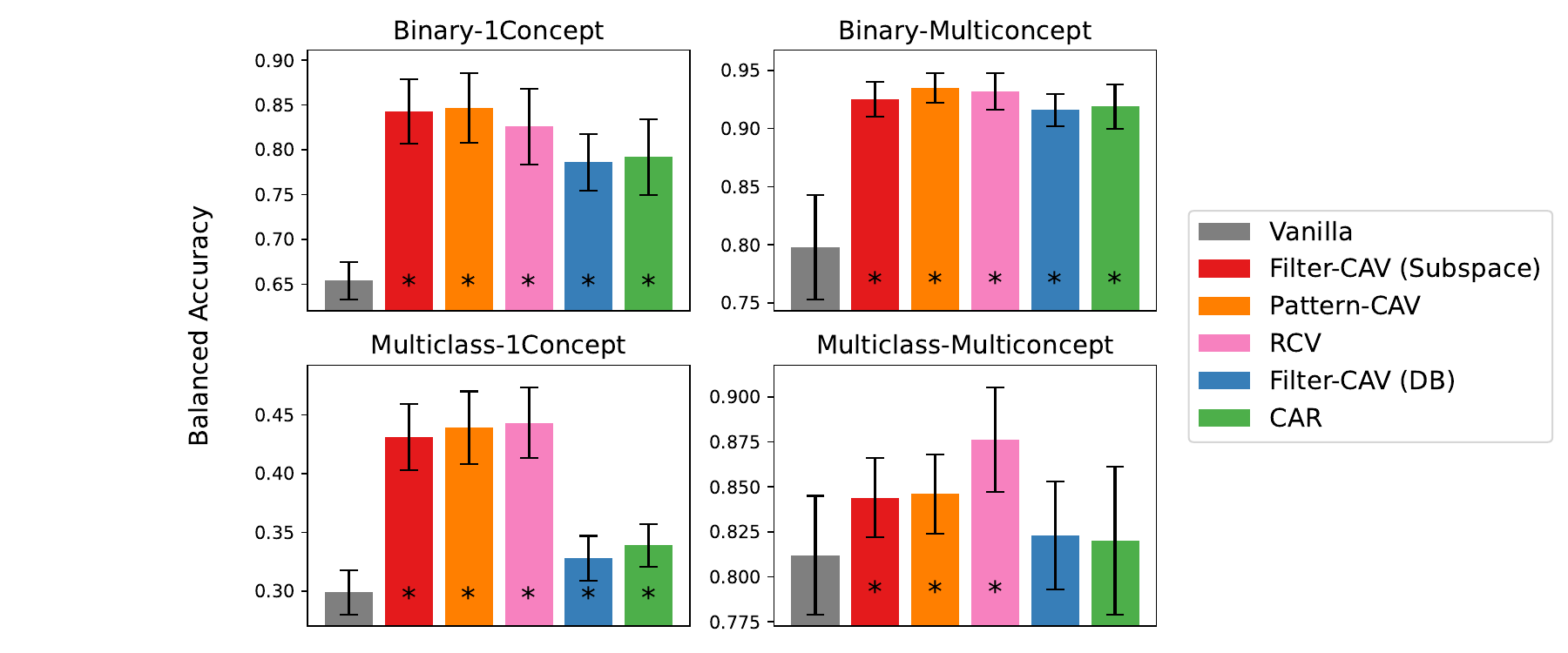}
    \caption{Balanced Accuracy (BA) of ResNet18 trained on a dataset that contains spurious correlations, and evaluated on a dataset without the correlation. Error bars indicate standard deviation over 20 runs, all on differently generated datasets. * indicates that the difference with the vanilla model is statistically significant at the 0.01 level, using a pairwise t-test.}
    \label{fig:res_elements}
\end{figure}

We extend our evaluation to assess robustness in more complex scenarios:

\begin{itemize}
    \item Binary-multiconcept: detect two shapes simultaneously (squares and triangles).
    \item Multiclass-1concept: classify images based on the number of squares (classes 0-4).
    \item Multiclass-multiconcept: classify images based on the presence of squares, triangles and circles. 
\end{itemize}

Because static regularisation uses the fewest hyperparameters and, as our optimisation search showed, performs on par with more complex setups, we employ \gls{lcrreg} in this mode. Moreover, we employ the hyperparameters tuned on the Binary-1Concept task for all the settings. From Figure~\ref{fig:res_elements}, we notice that \gls{lcrreg} consistently outperforms the vanilla baseline (ResNet18 without \gls{lcrreg}) across all settings. Cosine-based losses (Pattern-\gls{cav}, \gls{rcv}) yield statistically significant improvements. \gls{db}-based methods are effective but slightly less so. 

\subsection{Diabetic Retinopathy Classification}


\begin{figure}[t]
    \centering
    \includegraphics[width=0.95\linewidth]{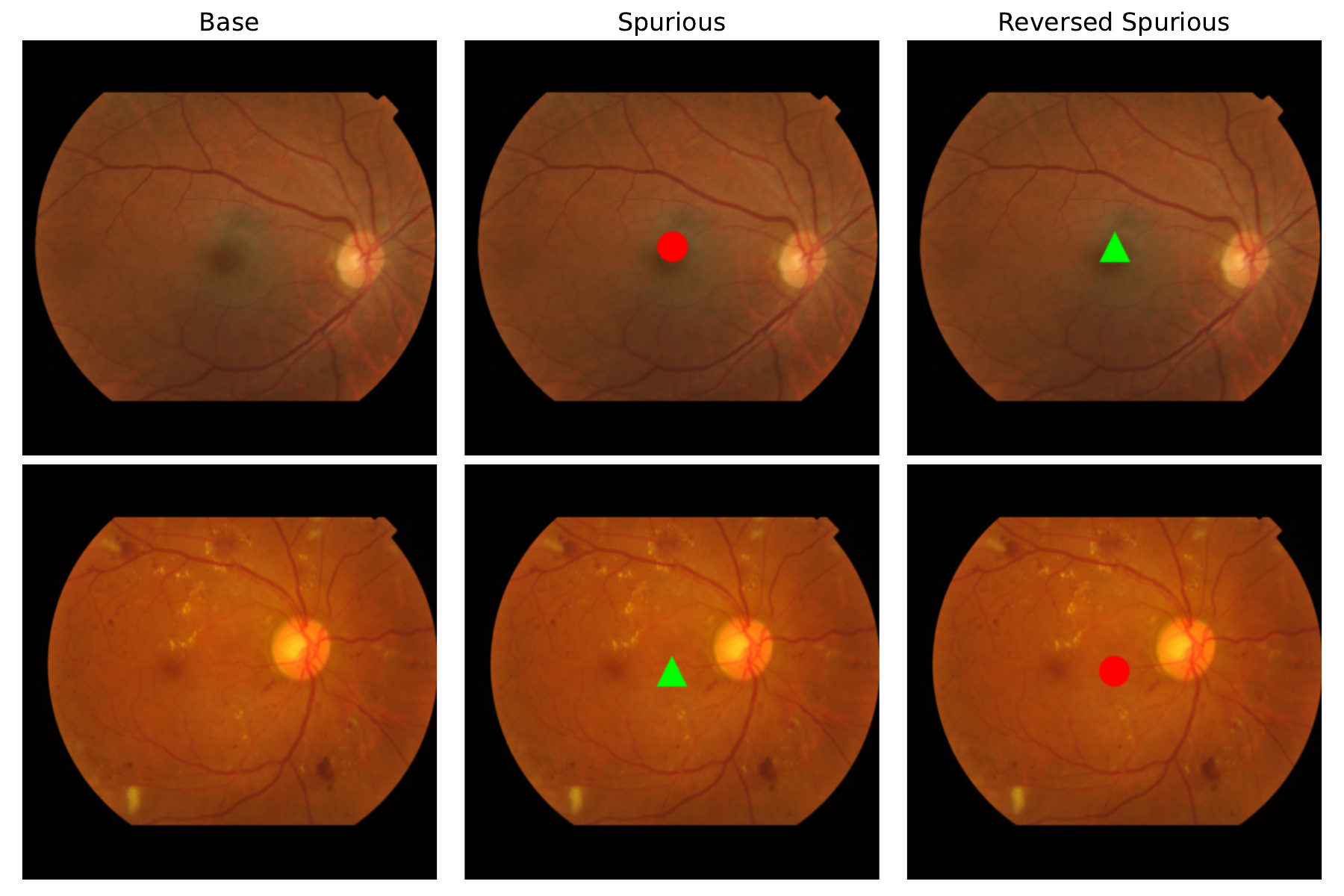}
    \caption{Example of spurious correlations on non-preprocessed images from APTOS. From left to right: the base images, with Spurious Correlations, with the Reversed Spurious Correlations.}
    \label{fig:spur_db}
\end{figure}

We evaluate \gls{lcrreg} on the clinical task of \gls{dr} detection. Training uses the APTOS dataset~\cite{aptos2019-blindness-detection}, with the FGADR dataset~\cite{Zhou_2021} providing pixel-level annotations of microaneurysms, a concept signifying the arising of \gls{dr}. To create the concepts for \gls{dr} we follow the pipeline described in Section~\ref{sec:meth}; example figures and more details can be found in Appendix~\ref{app:dr_concept_creation}. We evaluate model robustness against artificially injected spurious correlations (synthetic visual markers, shown in Figure~\ref{fig:spur_db}) and generalisation to \gls{ood} datasets.

\subsubsection{Robustness to Spurious Correlations}\label{subsub:rob}

\begin{table*}[t]
    \small
    \centering
    \caption{Out-of-Distribution performance in Balanced Accuracy (BA) (std) for vanilla and LCRReg models. Models were trained on APTOS via the same protocol, and evaluated on the entirety OOD datasets. Results are averaged over 5 runs.}
    \begin{tabular}{lc|cccccc}
    %
    \toprule
    &\textbf{In-Distribution}& \multicolumn{6}{c}{\textbf{Out-Of-Distribution}}\\
    \textbf{Model} & \textbf{APTOS} & \textbf{RLDR} & \textbf{IDRiD} & \textbf{DEEPDR} & \textbf{DDR}  & \textbf{EYEPACS} & \textbf{Avg}\\
    \midrule
    Vanilla & 96.8 (2.1) & 52.06 (1.83) & 68.67 (6.46) & 55.83 (2.60) & 57.17 (3.96) & 52.10 (0.01) & 57.17\\
    +LCRReg & 94.7 (2.4) & 51.75 (1.36) & 69.48 (3.54) & 58.33 (2.48) & 59.70 (1.77) & 52.31 (5.20) & 58.31 \\
    \midrule
    PCBM-h   & 96.4 (1.8) & 54.14 (2.10) & 69.04 (6.03) & 62.26 (3.72) & 60.38 (3.81) & 53.08 (0.01) & 59.78\\
    + LCRReg & 95.2 (3.1) & 54.58 (2.26) & 71.85 (8.80) & 62.85 (4.08) & 61.84 (2.85) & 54.79 (1.68) & 61.18\\
    \bottomrule
    \label{tab:ood}
    \end{tabular}
    \label{tab:ood_performance}
\end{table*}

We formulate \gls{dr} as a binary classification (healthy vs. unhealthy) and artificially inject strong synthetic markers correlated with class labels during training. We evaluate robustness to three scenarios:

\begin{itemize}
    \item Spurious dataset: the test set retains original spurious correlation distribution.
    \item Reversed spurious dataset: the spurious correlations are inverted (markers switched classes).
    \item Base dataset: original test set, without the artificial correlations.
\end{itemize}

\paragraph{Baselines.}

To contextualise our results, we compare \gls{lcrreg} against several baselines:

\begin{itemize}
    \item \gls{pcbmh}~\cite{yuksekgonul2022post} projects activations onto a concept subspace defined by post-hoc \glspl{cav} and employs residual fitting to preserve accuracy. \gls{pcbmh} provides interpretability through concept alignment but does not explicitly regularise against spurious correlations. 
    \item \gls{mtl} jointly trains the model on the primary classification task and an auxiliary task predicting concept presence. By explicitly supervising concepts during training, \gls{mtl} may implicitly encourage reliance on clinically meaningful features. 
    \item Linear probing is similar to \gls{mtl}, but instead of placing the prediction head on top of the final features, it is attached to the intermediate ones, specifically in the same layer where \glspl{lcr} are computed. This allows us to assess whether the \gls{lcr} calculation pipeline is necessary or if we could simply add a linear classification layer on the features. 
\end{itemize}

\paragraph{Training Setup and Hyperparameter Tuning.}

\begin{figure}[t]
    \centering
    \includegraphics[width=\columnwidth]{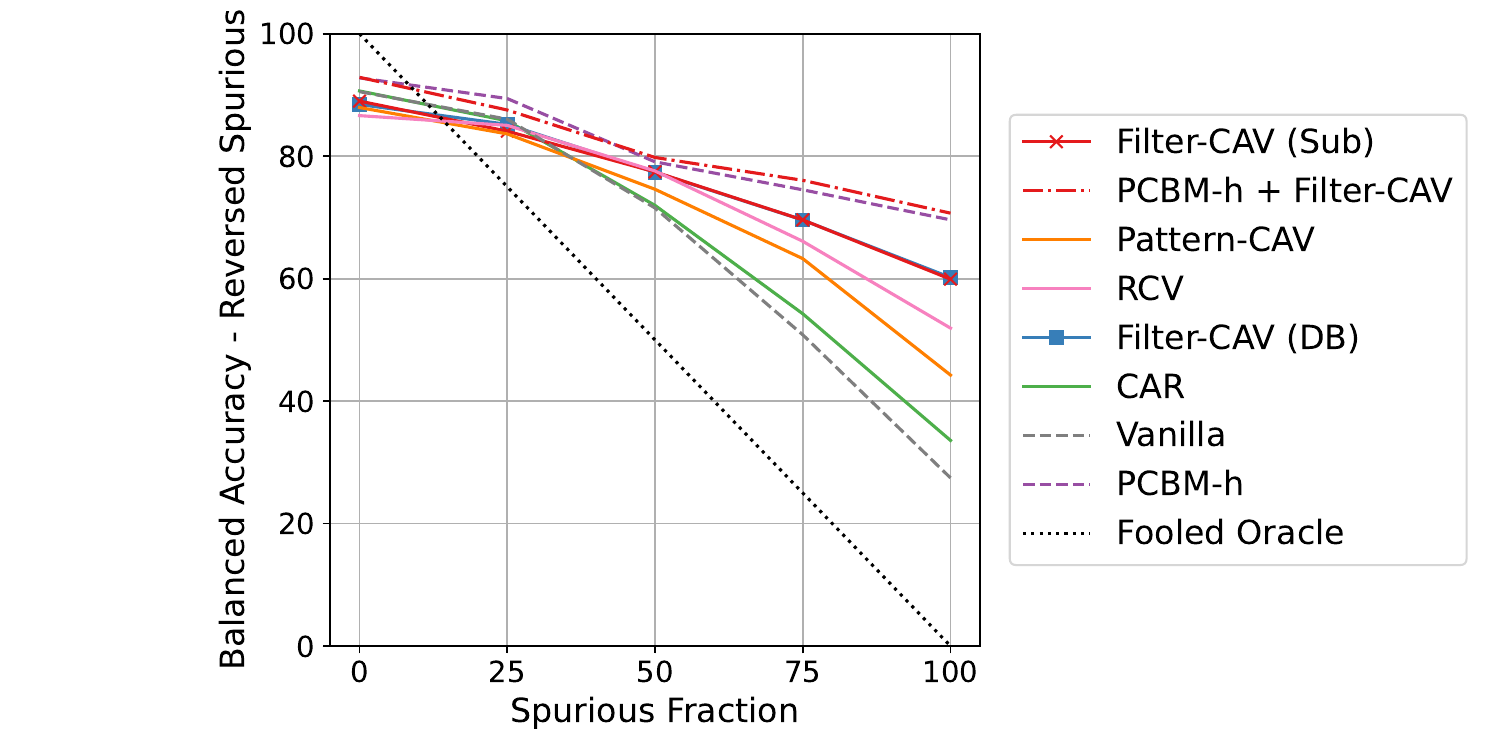}
    \caption{Performance averaged over 5 runs of the various LCRReg models, PCBM-h, and the vanilla model, assessed across multiple values of $p_{SC}$. Hyperparameters are tuned on the same data distribution of the training set.}
    \label{fig:res_spur_db}
\end{figure}

We evaluate models at different levels of spurious correlation presence, denoted as $p_{SC} \in \{0.00, 0.25, 0.50, 0.75, 1.00\}$, using a ResNet50~\cite{he2016deep} backbone, pretrained on ImageNet. 

Results are reported on a test set with reversed spurious correlations, averaged over 5 runs, and can be seen in detail in Appendix~\ref{app:spur_db}. Hyperparameters tuned via Optuna on validation sets matching the training distribution yield minimal robustness improvement, as optimisation favours exploiting spurious cues. Among these tuned models, only \gls{pcbmh} demonstrates consistent robustness across varying $p_{SC}$ levels, maintaining high predictive performance until spurious correlations dominate the entire training set ($p_{SC} = 1$).

Therefore, we select robust parameters derived from the synthetic experiments  ($\alpha_t = 100$, $\tilde{t} = 0$, $I_{rec} = \infty$). These settings substantially improved robustness particularly on the reversed spurious dataset (Figure~\ref{fig:res_spur_db}). \gls{lcrreg} significantly outperforms the baseline models, with Filter-\gls{cav} achieving the best overall performance. Combining \gls{lcrreg} with \gls{pcbmh} provides additional incremental robustness improvements, leveraging both regularisation during training and interpretability post-training. 

\begin{figure}
    \centering
    \includegraphics[width=0.95\linewidth]{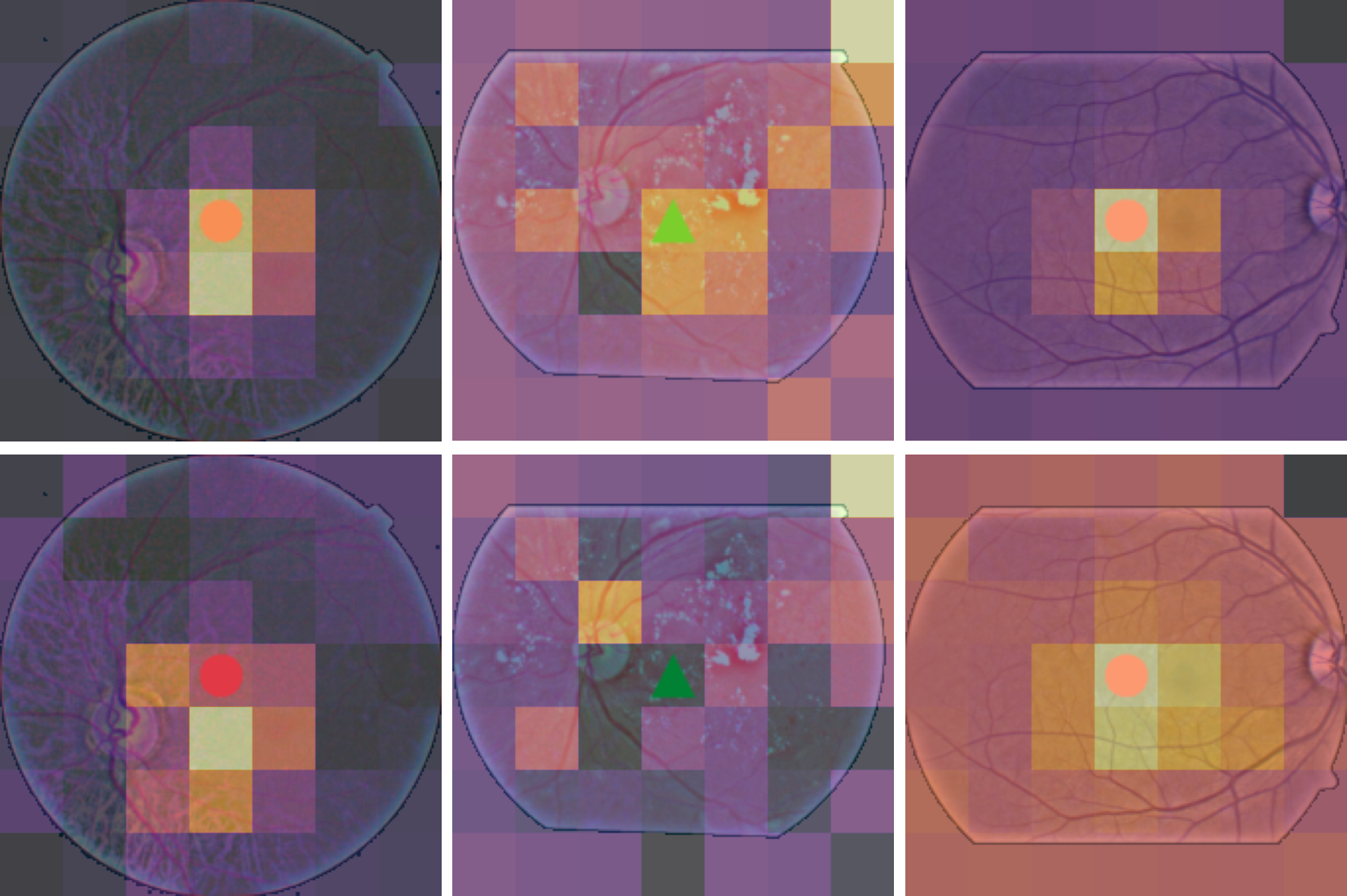}
    \caption{Example saliency map overlays from the vanilla model (top row) and the same model with LCRReg applied (bottom row). Each column corresponds to the same test image across models. Yellow indicates stronger activation; purple indicates weaker activation.}
    \label{fig:gradcam_dr}
\end{figure}

Saliency map visualisations depicted in Figure~\ref{fig:gradcam_dr} qualitatively confirm \gls{lcrreg} reduces reliance on spurious correlations compared to the vanilla baseline. 

\subsubsection{OOD Generalisation}

We assess generalisation by training on APTOS and evaluating on five unseen datsets: RLDR~\cite{rldr}, IDRiD~\cite{porwal2018indian}, DEEPDR~\cite{deepdr}, DDR~\cite{ddr}, and EYEPACS~\cite{eyepacs}. \gls{lcrreg} achieves modest yet statistically significant improvements over the vanilla baseline and \gls{pcbmh} (Wilcoxon test, $p = 0.0058$), as can be seen from Table~\ref{tab:ood_performance}.

\subsubsection{Ablation on Model Architectures}\label{subsub:abl_arch}

We evaluate \gls{lcrreg}'s generalisability across different backbone models. Results reported in Table~\ref{tab:model_ablation} in Appendix~\ref{app:ablations} confirm reduced reliance on spurious correlations, though the results are not statistically significant individually. Notably, the effect is minimal for ResNet152, suggesting that the current regularisation method is not yet sufficient to effectively address spurious correlations in larger architectures.

\section{Discussion} \label{sec:disc}

Our results demonstrate that \gls{lcrreg}, a plug-in regularisation method leveraging \glspl{lcr}, can improve robustness to spurious correlations in both synthetic and real-world medical image classification tasks. Compared to traditional methods requiring dense concept supervision, \gls{lcrreg} operates without concept labels on the main dataset, using instead a lightweight synthetic concept bank generated via compositional editing. This improves scalability for real-world deployment.

In synthetic settings, \gls{lcrreg} consistently improved robustness of the vanilla model across all experimental configurations. Performance gains were strongest in the Binary-1Concept task, where hyperparameters were tuned, but remained competitive even when the same parameters were reused in more complex multi-concept and multiclass settings. Notably, \gls{db}-based methods showed decreasing relative benefit as task complexity increased, suggesting that cosine-based subspace regularisation may generalise better in high-dimensional concept spaces.

In real-world experiments, \gls{lcrreg} outperformed standard baselines, like multitask learning and linear probing. While \gls{pcbmh} remained the strongest stand-alone baseline under distribution shift, combining it with \gls{lcrreg} offered modest additional robustness gains, highlighting the complementary strengths of regularisation during training and concept-based adjustments post hoc.

Among \gls{lcr} types, filter-\glspl{cav} provided the most reliable robustness gains in the clinical setting, despite being less sophisticated than \glspl{car} or pattern-\glspl{cav}. This aligns with the hypothesis that discriminative directions, rather than descriptive activations, offer a stronger inductive bias when used for regularisation. Frequent \gls{lcr} recomputation did not improve results; on the contrary, recomputing \glspl{cav} only once at the beginning yielded more stable and robust performance, likely due to avoiding moving-target dynamics during optimisation.

Hyperparameter tuning further revealed consistent trends: starting regularisation early and using a high regularisation weight ($\alpha_t$) improved performance, especially for cosine-based loss formulations. However, tuning remains non-trivial in the presence of spurious correlations, as optimising on in-distribution validation sets often favours learning spurious cues. Future methods may benefit from tuning procedures guided by robustness-aware metrics or proxy \gls{ood} validation splits.
\section{Conclusion}\label{sec:conc}

We introduce \gls{lcrreg}, a scalable regularisation method that uses \glspl{lcr} derived from synthetic concept images to guide model training toward semantically meaningful features. Our approach requires only minimal auxiliary supervision and can be easily integrated into existing architectures as a modular regularisation technique.

Experiments on both a controlled synthetic dataset and \gls{dr} classification show that \gls{lcrreg} improves robustness to spurious correlations and provides modest gains in \gls{ood} generalisation. It outperforms more complex concept supervision baselines under the same constraints and shows complementary benefits when combined with post hoc approaches such as \gls{pcbmh}.

Despite its simplicity, filter-\gls{cav} remains the most effective \gls{lcr} type, suggesting that separation-based concept vectors are particularly well-suited for regularisation. Our results also show that strong early regularisation with static concept vectors is a viable and stable training strategy. Surprisingly, we found that \gls{lcrreg} performed best when the \gls{cav} was computed only once. While internal representations evolve during training, updating the \glspl{cav} too frequently introduced instability, likely due to the moving target problem. A promising solution may involve using a moving average of the \gls{cav}, enabling smoother updates and more stable training. 

These findings underscore the potential of using concept-based signals for robustness without requiring extensive concept-label supervision. Future work may explore extending \gls{lcrreg} to deeper models, multi-concept settings, and combining it with gradient manipulation techniques (e.g., Gradient Surgery~\cite{mansilla2021domain}) to further improve training stability and scalability.
\section*{Acknowledgements}
Research at the Netherlands Cancer Institute is supported by grants from the Dutch Cancer Society and the Dutch Ministry of Health, Welfare and Sport.
Kristoffer Wickstrøm is financially supported by the Research Council of Norway, through its Center for Research-based Innovation funding scheme (grant no. 309439).
{
    \small
    \bibliographystyle{ieeenat_fullname}
    \bibliography{main}
}

\clearpage
\appendix
\section{Diabetic Retinopathy Concept Creation}
\label{app:dr_concept_creation}
\begin{figure}[htb]
    \centering
    \includegraphics[width=\linewidth]{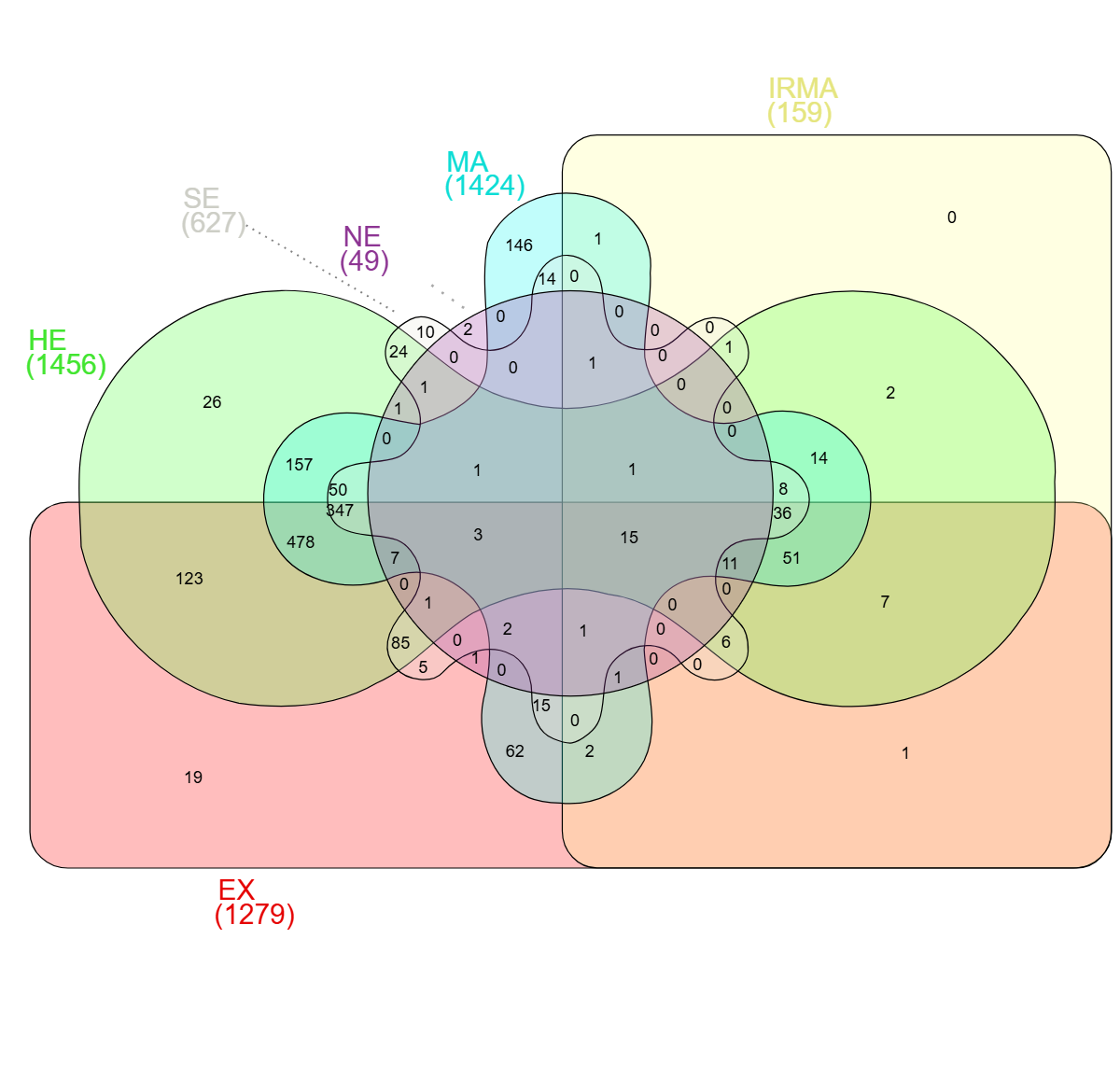} 
    \caption{Concept entanglement diagram for FGADR dataset, showcasing how often all concepts coincide in a single image.}
    \label{fig:concept-overlap}
\end{figure}

For creating images representing diagnostically relevant concepts, we use the pixel-wise annotations of the FGADR dataset to construct positive and negative concepts. An illustration of concept overlap in the fundus images, motivating the need for a controlled creation pipeline, is provided in Figure~\ref{fig:concept-overlap}.

First, we horizontally flip all left-eye images to ensure consistent orientation across the dataset, so that all eyes appear as right-eye views. This will not introduce bias, as all training images can be randomly flipped. 

For creating backgrounds, we start with defining the set of healthy images. We take all images that do not have any lesion annotations. Then, a subjective visual assessment is made to exclude healthy images with irregular patterns. Although possibly introducing bias, this is done to ensure the remaining candidates can serve as a background that does not introduce much noise. 

\begin{figure}[htb]
    \centering
    \includegraphics[width=0.95\linewidth]{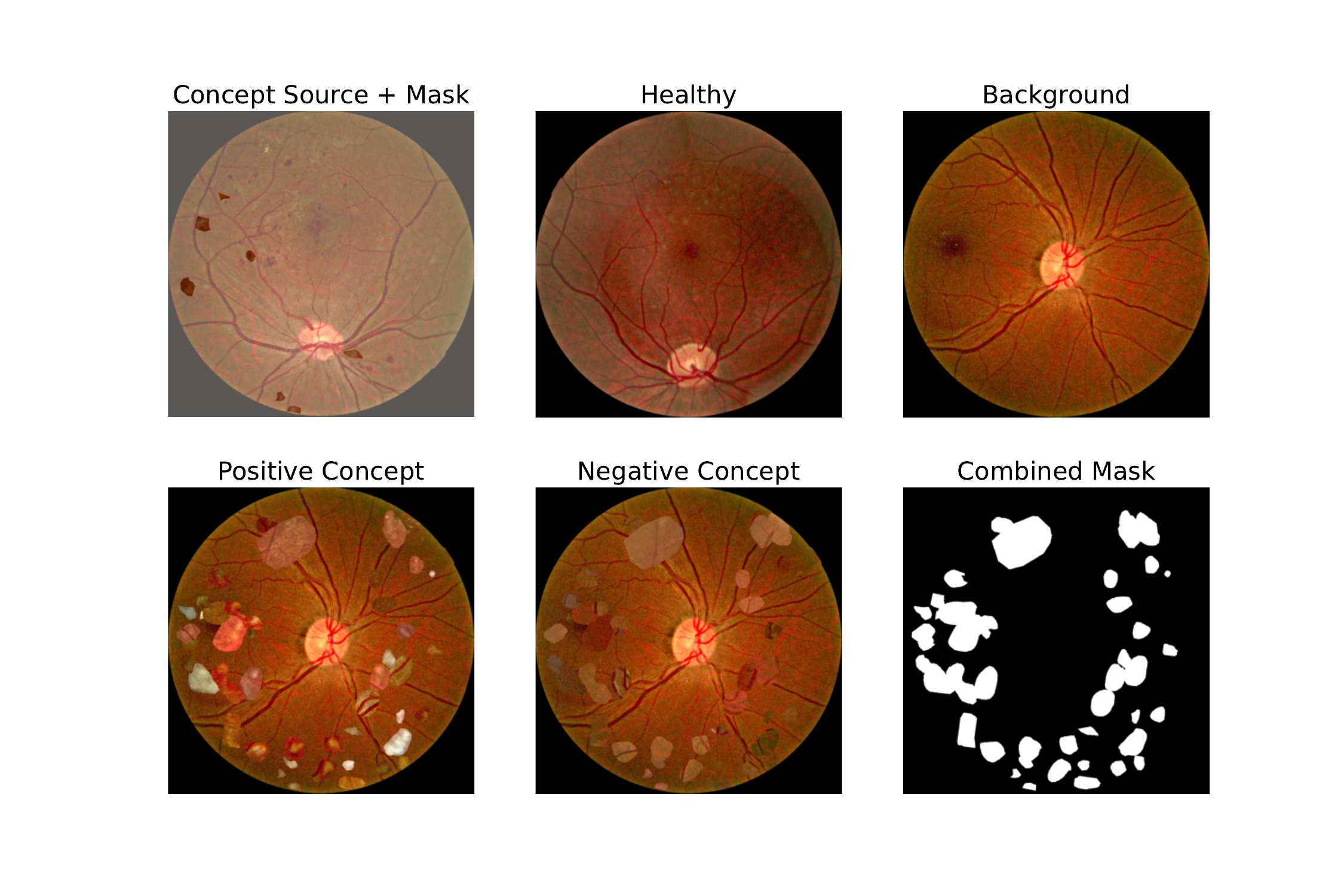}
    \caption{Example of concept creation pipeline on the fundus images from the FGADR dataset for concept: \emph{Soft Exudates}. Done on non-preprocessed data for visualisation, as patches can be seen more easily without correcting for colour.}
    \label{fig:FGADR_concept_example}
\end{figure}

After these steps, we apply the concept creation pipeline to create 128 positive and negative concept image pairs. This was a trade-off decision: while more concepts would always be preferable for creating stable \glspl{lcr}, increasing the number would require reusing source images more often, reducing variability across concept examples. Note that the number of 128 could be sufficient, as natural images only need a couple dozen images~\citep{kim2018interpretability}. An example of such a pair is shown in Figure~\ref{fig:FGADR_concept_example}.
\paragraph{Internal Consistency and Domain Robustness.}


\begin{figure*}[htb]
  \centering
  \begin{subfigure}[b]{0.48\textwidth}
    \centering
    \includegraphics[width=\linewidth]{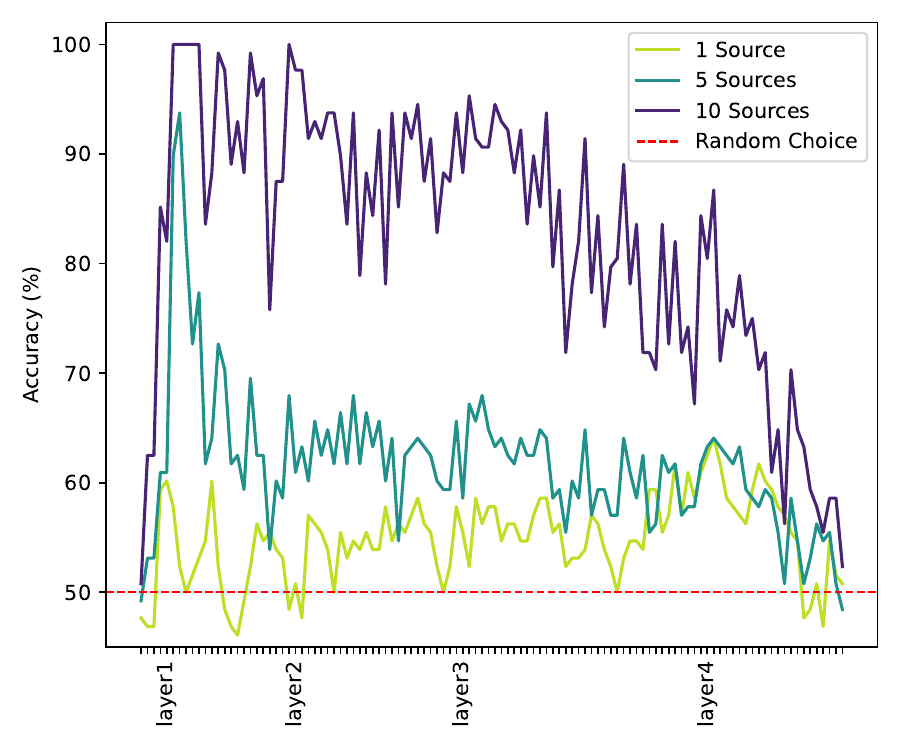}
    \caption{}
  \end{subfigure}
  \hfill
  \begin{subfigure}[b]{0.48\textwidth}
    \centering
    \includegraphics[width=\linewidth]{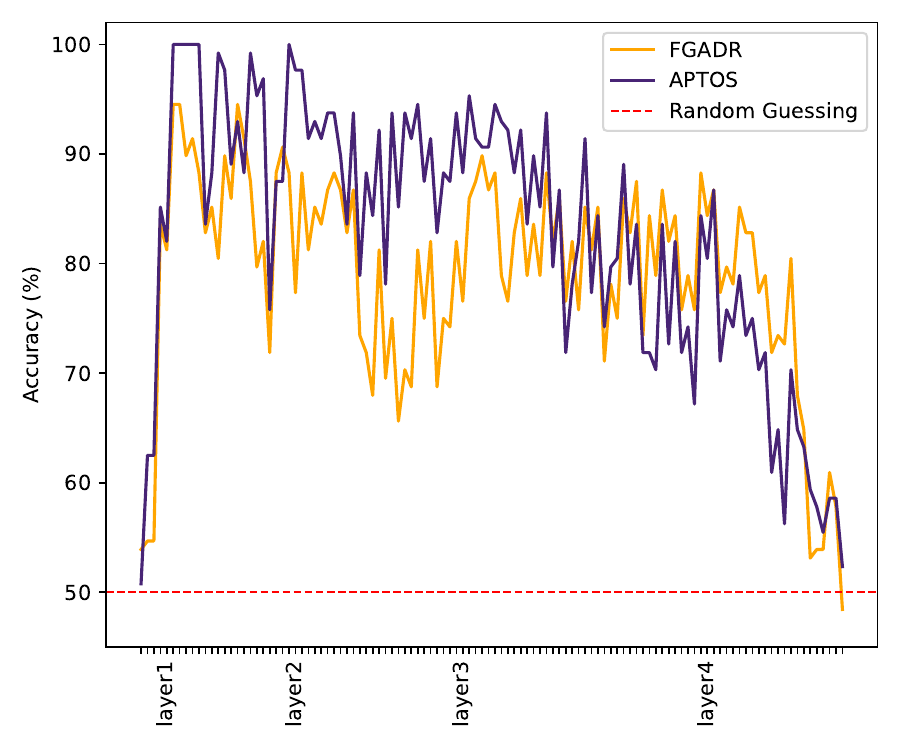}
    \caption{}
  \end{subfigure}
  \caption{Illustrations of the internal coherence score per layer of a ResNet50 trained on APTOS. Filter-CAV was calculated using 128 images, and evaluated using 64 other images, using accuracy of the internal decision boundary. The train-test split was created using different source images. (a) shows the relation between number of source images per concept example, while (b) depicts the effect of changing the models' training data.}
  \label{fig:concpet_creation_experiments}
\end{figure*}

We conducted experiments to evaluate whether using multiple source images per concept pair improves the quality of the learnt \glspl{lcr}. Results of these experiments are shown in Figure~\ref{fig:concpet_creation_experiments}. To this end, we measured the accuracy of the internal \gls{db} of filter-\glspl{cav}, computed across all layers of a ResNet50 trained on the APTOS dataset.

While high \gls{db} scores indicate that positive and negative concept examples are well-separated in the latent space, this alone does not confirm that the resulting \glspl{lcr} are meaningful. Rather, it provides a necessary, but not sufficient, condition for their utility in \gls{lcrreg}. If the concepts were not consistently separable, the resulting \glspl{db} and \glspl{cav} would be no more informative than random directions in the representation space.

In addition, we evaluated the \glspl{cav} on a model trained on the FGADR dataset, the same dataset from which the concepts were derived. This experiment was conducted to assess whether concept performance improves significantly when evaluated in-domain. If performance on the FGADR-trained model is not substantially higher than on a model trained on APTOS, this suggests that the concepts are robust to data shift, as their utility does not strongly depend on the training domain of the model.

\paragraph{Results.} We found that using only a single source image per concept pair resulted in near-random \gls{db} scores, indicating poor separability in the latent space. Increasing the number of source images improved performance: using five images led to substantially better results, and using ten yielded further improvements. We did not extend beyond ten images due to limited data availability and the need to preserve variability across concept examples. \gls{db} accuracies were comparable between models trained on APTOS and FGADR, suggesting that the learnt concepts are robust to data shift and do not rely heavily on the training domain.

\section{Ablations}
\label{app:ablations}

\begin{figure*}[htb]
    \centering
    \begin{subfigure}[b]{0.32\textwidth}
        \centering
        \includegraphics[width=\textwidth]{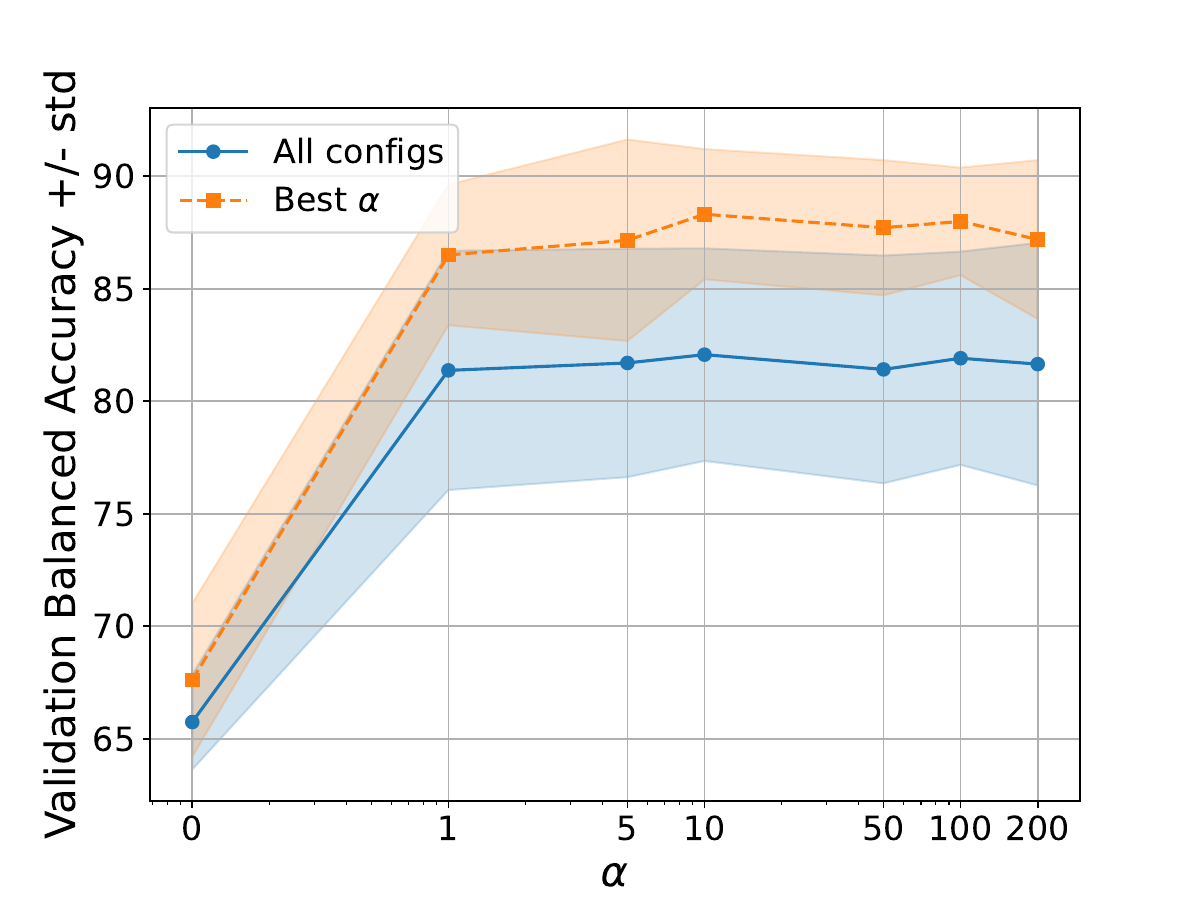}
        \caption{}
        \label{fig:varying_alpha}
    \end{subfigure}
    \hfill
    \begin{subfigure}[b]{0.32\textwidth}
        \centering
        \includegraphics[width=\textwidth]{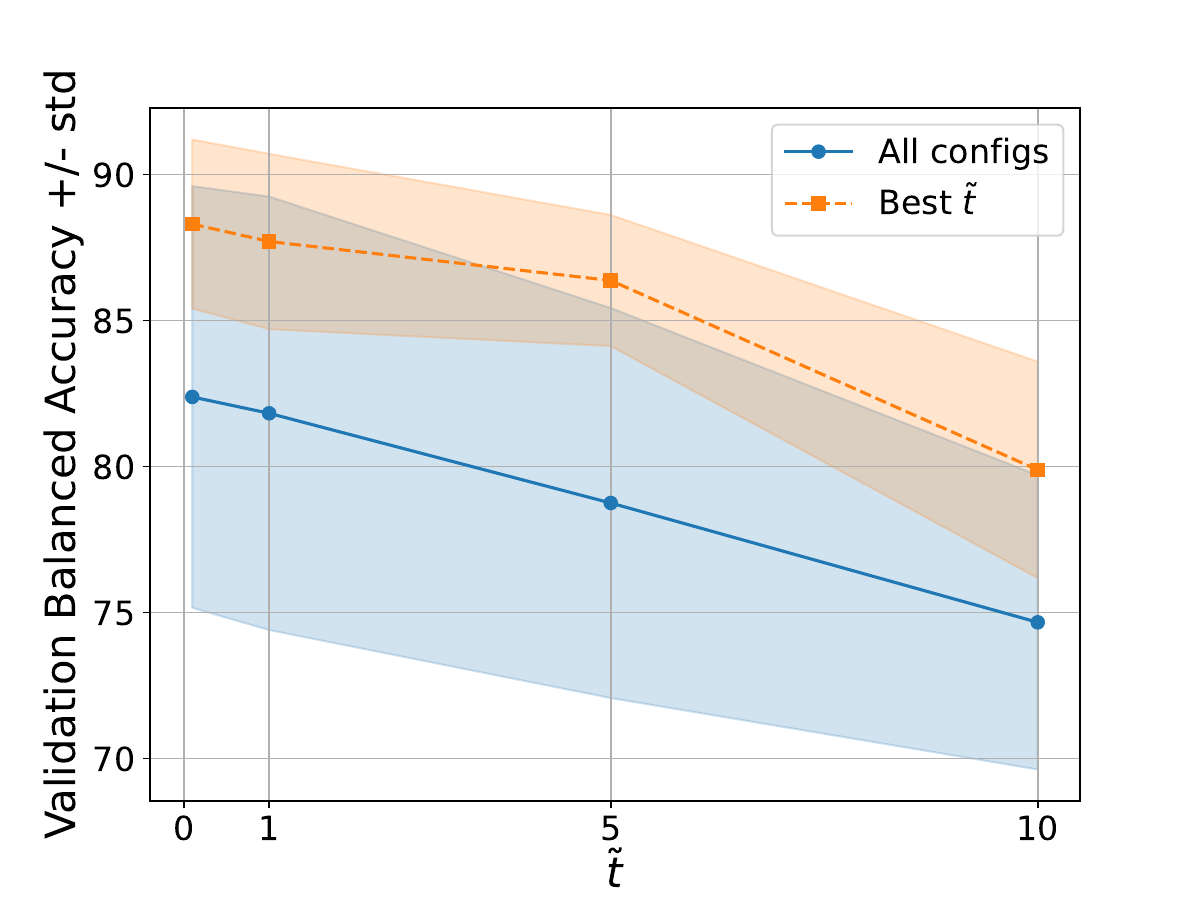}
        \caption{}
        \label{fig:varying_start}
    \end{subfigure}
    \hfill
    \begin{subfigure}[b]{0.32\textwidth}
        \centering
        \includegraphics[width=\textwidth]{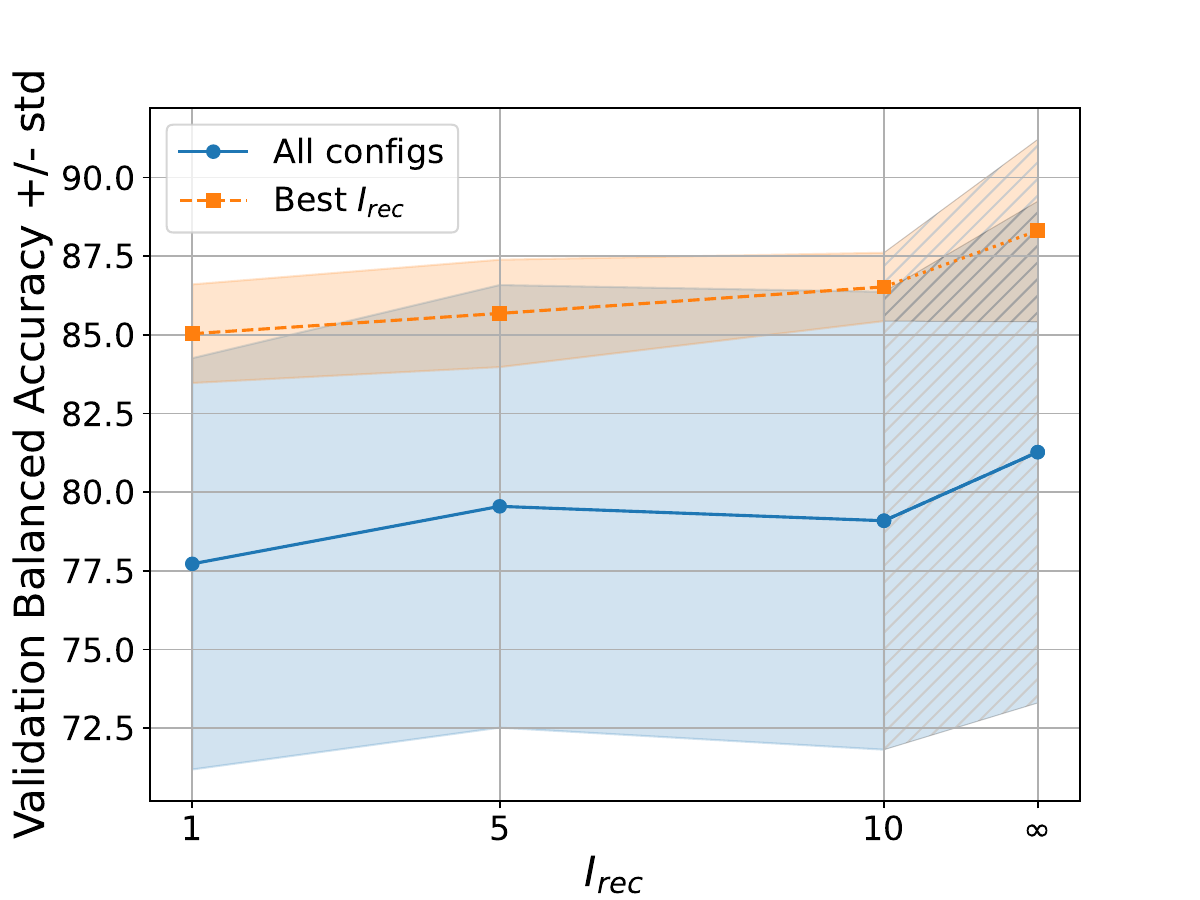}
        \caption{}
        \label{fig:varying_recomp}
    \end{subfigure}
        \caption{Results of hyperparameter grid search on the binary classification task (Elements dataset) using filter-CAV with cosine loss. Each configuration was run five times with different train/validation splits. Blue lines: mean $\pm$ std over all combinations of remaining parameters. Orange lines: mean $\pm$ std with best setting of other parameters per fixed value. (a) Weight of the $\mathcal{L}_{LCRReg}$: $\alpha_t$. (b) Starting epoch: $\tilde t$. (c) Recomputation interval: $I_{rec}$.}
    \label{fig:hyperparameter_ablation}
\end{figure*}

\begin{figure*}[htb]
  \centering
  \begin{subfigure}[b]{0.48\textwidth}
    \centering
    \includegraphics[width=\linewidth]{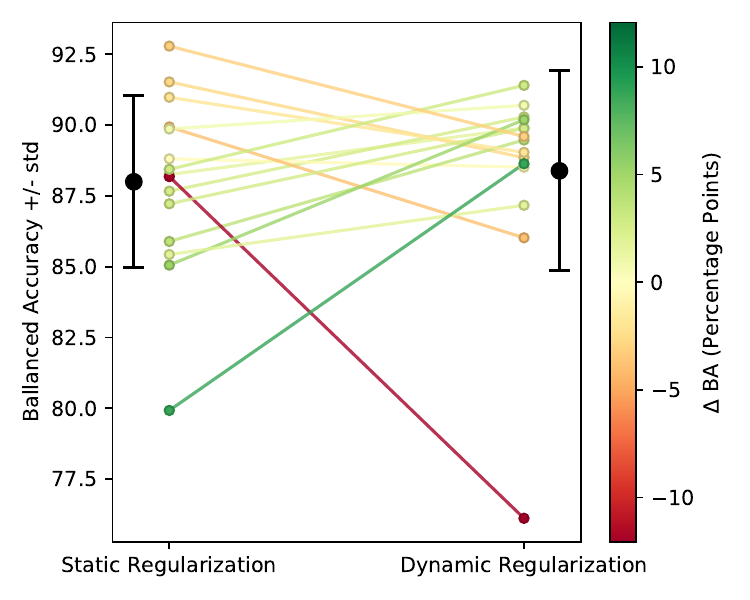}
    \caption{}
    \label{fig:statvsdyn}
  \end{subfigure}
  \hfill
  \begin{subfigure}[b]{0.48\textwidth}
    \centering
    \includegraphics[width=\linewidth]{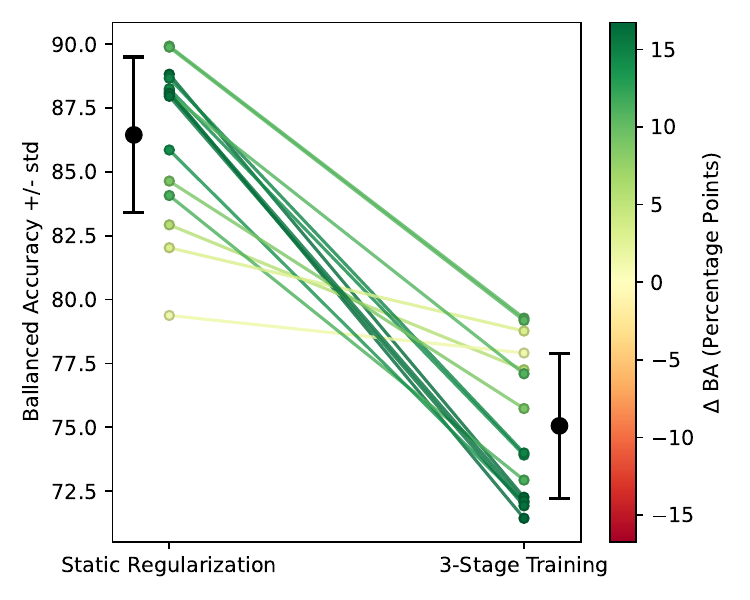}
    \caption{}
    \label{fig:statvs3st}
  \end{subfigure}
  \caption{Comparison of different Static regularisation to both Dynamic regularisation, and 3–Stage Training, evaluating all approaches on 15 distinct train–test dataset pairs. The figure shows both the Balanced Accuracy (BA) of individual runs, and the total average. Statistical comparison was done via pairwise t-testing. Comparison of Static regularisation vs. (a) Dynamic regularisation and vs. (b) 3–Stage Training.}
  \label{fig:reg_scheme_comp}
\end{figure*}

In this Appendix, we report further results of ablations studies. In particular,~\Cref{fig:hyperparameter_ablation} and~\Cref{fig:reg_scheme_comp} show additional ablations on hyperparameters and training strategies on the Elements dataset, discussed in~\Cref{subsec:elem}.~\Cref{tab:model_ablation} shows performance of LCRReg on different model architectures, which is addressed in~\Cref{subsub:abl_arch}.

\begin{figure}[!t]
    \centering
    \includegraphics[width=\linewidth]{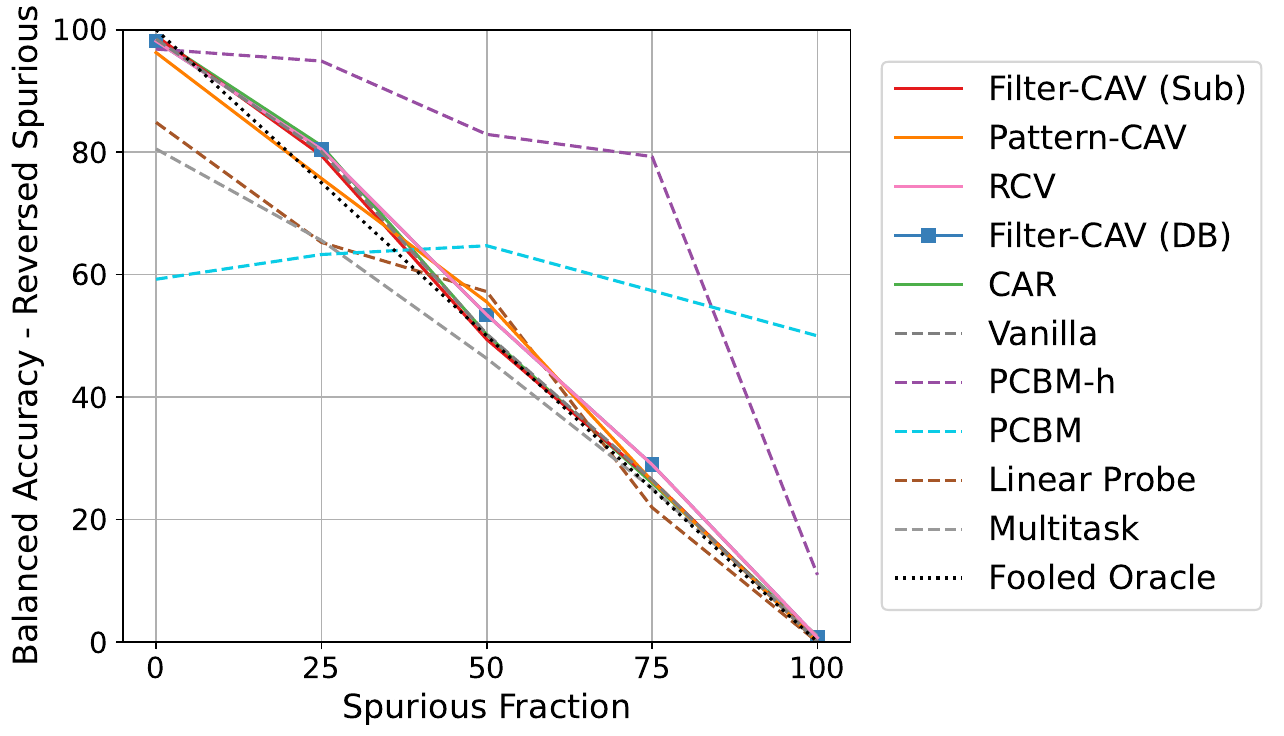}
    \caption{Performance averaged over 5 runs of various LCRReg models, PCBM-h, and the vanilla model, assessed across multiple values of $p_{SC}$. Models are using hyperparameters that are tuned using Optuna, with the validation set being of similar distribution as the training set.}
    \label{fig:tuned_performance_aptos}
\end{figure}

\begin{table*}[t]
\centering
\caption{Performance of Latent Concept Representation-based regularisation across different datasets and model architectures, trained and evaluated on APTOS. Trained on a Spurious Dataset with $p_{SC}=1$. Performance reported in terms of balanced accuracy. Parameter counts (in millions) are shown in parentheses below model names.}
\begin{tabular}{ll|cccc}
\toprule
 \textbf{Dataset} &  \textbf{Model} & \textbf{ResNet18} & \textbf{ResNet152} & \textbf{DenseNet121} & \textbf{InceptionV3} \\
&&(11.6 M) &(60.2 M)&(8.0 M)& (23.9 M)\\
\midrule
 Reverse Spurious & Vanilla   & 14.81 (9.87) & 14.60 (9.76) & 43.71 (13.10) & 17.96 (4.32) \\
& Regularised & 31.47 (15.55) & 16.82 (16.07) & 65.74 (11.83) & 49.12 (6.27) \\
\midrule

 Base & Vanilla     & 75.32 (5.70) & 84.81 (6.16) & 88.91 (4.15) & 79.66 (6.16) \\
& Regularised & 80.14 (5.72) & 82.32 (1.56) & 85.26 (2.42) & 66.25 (9.80) \\
\midrule

 Spurious & Vanilla     & 99.69 (0.23) & 99.66 (0.26) & 99.06 (0.50) & 97.22 (1.47) \\
& Regularised & 99.49 (0.30) & 99.32 (0.86) & 93.78 (3.58) & 80.49 (12.0) \\

\bottomrule
\end{tabular}
\label{tab:model_ablation}
\end{table*}


\section{Robustness to Spurious Correlations}
\label{app:spur_db}
\Cref{fig:tuned_performance_aptos} reports the results of the different models on the APTOS dataset with the hyperparameters finetuned with Optuna, commented in~\Cref{subsub:rob}. 
Note that we also provide the accuracy of \gls{pcbm}, which is equal to \gls{pcbmh} without residual fitting. However, this is not discussed in the main section due to it not achieving performance significantly different than random performance, and it is thus not reported in~\Cref{fig:res_spur_db} in the Main Text.

\end{document}